\documentclass{dhbenelux}

\usepackage{booktabs} 
\usepackage{authblk} 
\usepackage{graphicx} 
\usepackage{amssymb}
\usepackage{bbding}
\usepackage{utfsym}
\usepackage{hyperref}

\usepackage{indentfirst}
\author[,a,b]{Chao Huang\textsuperscript{1,2}}
\author[,b]{Chunyan Chen\textsuperscript{1}}
\author[,b]{Ling Shi\textsuperscript{1}}
\author[b]{Chen Chen}

\affil[a]{Institute of Computing Technology, Chinese Academy of Science, Beijing, China}
\affil[b]{Ningbo Institute of Information Technology Application, Chinese Academy of Sciences (CAS), Ningbo, China}

\title{Material Property Prediction with Element Attribute Knowledge Graphs and Multimodal Representation Learning}

\begin{document}

\maketitle
\footnotetext[1]{These authors have made equal contributions}
\footnotetext[2]{Corresponding author: chuang@ict.ac.cn}

\begin{abstract}
    \noindent {Machine learning has become a crucial tool for predicting the properties of crystalline materials. However, existing methods primarily represent material information by constructing multi-edge graphs of crystal structures, often overlooking the chemical and physical properties of elements (such as atomic radius, electronegativity, melting point, and ionization energy), which have a significant impact on material performance. To address this limitation, we first constructed an element property knowledge graph and utilized an embedding model to encode the element attributes within the knowledge graph. Furthermore, we propose a multimodal fusion framework, ESNet, which integrates element property features with crystal structure features to generate joint multimodal representations. This provides a more comprehensive perspective for predicting the performance of crystalline materials, enabling the model to consider both microstructural composition and chemical characteristics of the materials. We conducted experiments on the Materials Project benchmark dataset, which showed leading performance in the bandgap prediction task and achieved results on a par with existing benchmarks in the formation energy prediction task.}
\end{abstract}


\section{Introduction}

The prediction of material properties is an important aspect of materials engineering applications, such as the discovery of novel materials\cite{hamilton2024machine} with specific properties and the assessment of the reliability of materials in use\cite{zhao2024machine}. Computational methods based on quantum mechanics (e.g. density functional theory, DFT) play a key role in predicting the physical and chemical properties of materials, but the high computational complexity, high cost and long computation time of such methods severely limit their applicability to large-scale materials systems. In recent years, machine learning methods have been adopted by an increasing number of research institutes due to their potential to efficiently and accurately predict material properties, opening up new avenues for rapid screening and optimisation of materials\cite{mueller2016machine,kong2021materials,zhang2023artificial,hwang2023stability,Banik2024EvaluatingGF}. In these machine learning models, the characterisation of the material's crystal structure is crucial.

The crystal structure of a material is usually modelled by the smallest cell containing all the constituent atoms in different coordinates, repeated infinitely many times in 3D space on a regular lattice, making the material structure periodic in nature. By modelling the material crystal structure with Graph Neural Networks (GNN), representations based on geometrical structural information have been constructed to enable prediction of material properties\cite{xie2018crystal,schutt2017schnet,choudhary2021atomistic,yan2024complete,Xie2018CrystalGC,materalsproject,louis2020graph,Choudhary2021AtomisticLG,alignn,Isayev2017UniversalFD,Yan2024CompleteAE}. These research methods have introduced various strategies such as the introduction of geometrical features such as multi-scale information, symmetry features, bond lengths and bond angles to improve the prediction accuracy of various crystalline properties. Das K et al.\cite{das2023crysmmnet} fused the multimodal features of the crystal structure and the textual representation of the structure to capture both the local domain features in the crystal structure and the global chemical features in the textual representation for the prediction of crystalline material properties. Although the method introduces multimodal data, the quality and consistency of the textual description cannot be guaranteed and must be generated using Robocrystallographer\cite{ganose2019robocrystallographer}, making the data pre-processing process complex and dependent. Although these studies can effectively model crystal structures and their inherent properties, existing methods are purely data-driven, focusing on exploring the intrinsic topological structure and structural rules of crystal structures without incorporating any chemical prior knowledge. The lack of this crucial elemental-level information often leads to inaccurate performance predictions when these models are applied to complex material systems containing different types of elements with significant variations in their properties.

In this study, we propose a multimodal fusion framework, ESNet, to enhance the accuracy of crystalline material property predictions by integrating element attributes with crystal structure features. First, we construct an element attribute knowledge graph that systematically captures the chemical and physical properties of elements (such as electronegativity, atomic energy, and modulus). Using an embedding model to encode this knowledge graph, we reveal deeper relationships among various elements, yielding a rich representation of elemental features. Then, ESNet jointly learns both element attribute and crystal structure features, allowing the model to represent material microcomposition and chemical characteristics from a more comprehensive perspective. For the extraction of crystal structure features, we reference the ComFormer\cite{Yan2024CompleteAE} approach. Extensive experiments on the Materials Project benchmark dataset validate the effectiveness and superiority of ESNet in predicting key material properties such as band gap and formation energy. Results indicate that the multimodal representation combining element attributes and crystal structure significantly improves model performance, offering a novel approach to crystalline material property prediction.

\section{Background And Related Works}

\subsection{Crystal Structures}

The crystal structure of a material consists of a basic crystal cell and a set of atomic primitives associated with that cell, and its uniqueness is reflected in its periodicity and infinity. In three dimensions, the crystal structure is represented by the periodic translation of the crystal cell through a specific lattice matrix, which at the same time embodies infinite repeatability. This property of the crystal structure determines its physical properties, such as the symmetry and mechanical properties of the crystal, etc. Different crystal structures lead to different interactions between atoms, which in turn affect different physicochemical properties of the crystal.

In the mathematical representation of crystal structures, a common method is described by M=(A,P,L)\cite{yan2022periodic}, where A contains the eigenvectors of the n atoms in the crystal cell, P denotes the three-dimensional Euclidean positions of the atoms in the crystal cell, and L describes the repetitive pattern of the crystal cell in three-dimensional space. This representation is capable of accurately depicting the basic elements of the crystal structure and provides a fundamental framework for subsequent studies of crystal properties.

\subsection{Crystal property prediction}

The use of deep learning for crystal property prediction has become increasingly popular in recent years, as it offers computationally efficient alternatives to classical simulation methods. In the current research frontier of crystal property prediction, the results mainly focus on the relevant variants of the Transformer structure, which fully take into account the unique physical properties of the crystal structure, such as periodicity, SE(3) invariance, SO(3) isotropy, and other key attributes. Comformer\cite{yan2024complete}, by capturing the periodicity feature of the crystal structure, proposes a graph-based Transformer model, as well as SE(3)-invariant and SO(3)-isotropic crystal graph representations for geometrical integrity and effective prediction of crystals. Efficient message passing with iComFormer and eComFormer is used to ensure the computational complexity while fully taking into account the geometrical information to achieve accurate prediction of crystal material properties. MatFormer\cite{yan2022periodic} focuses on solving the problem of encoding the periodicity of the crystal structure by designing a fully connected graph construction method, which ensures that the model is unaffected by the periodic boundary variations. At the same time, self-connected edges are used to encode periodic patterns, and a special message-passing scheme is proposed to effectively update the node characteristics to improve the accuracy of crystal material property prediction.

\subsection{Knowledge Graph construction}

A knowledge graph, as an efficient knowledge organization model, is represented as a semantic network consisting of nodes and edges, where nodes represent entities in the physical world and edges represent properties or relationships of entities. Knowledge graphs were first proposed to improve search. With the development of artificial intelligence, a new trend is emerging, from data-driven intelligence to knowledge-data hybrid-driven intelligence, in which knowledge graphs play an increasingly important role. At present, the application areas of knowledge graphs are becoming more and more extensive, allowing the construction of knowledge graphs that belong to specific domains.

In the field of materials science, research on knowledge graphs is still in its infancy, and only a few researchers have explored the extraction of materials knowledge and conducted research on the extraction of generic types of knowledge such as material entities and relationships. In recent years, the construction and application of knowledge graphs in the field of materials is gradually becoming a research hotspot, and at the same time, significant results have been achieved, such as Vineeth Venugopal et al.\cite{venugopal2022matkg} constructed the largest knowledge graph in the field of materials science, MatKG; Jamie P. McCusker et al.\cite{mccusker2020nanomine} proposed a knowledge graph focused on nanocomposites science, NanoMine; and a knowledge graph based on the Knowledge Graph for High-Throughput Computing to study the correlation between the organisation and mechanical properties of 6XXX aluminium alloys\cite{zheng2022high}, etc.

It is worth noting that methods based on elemental knowledge graphs have already been proposed at the molecular level, and the KCL\cite{fang2022molecular} and KANO\cite{fang2023knowledge} frameworks are the first to apply chemical elemental knowledge graphs in molecular comparative learning, which are capable of exploiting chemical fundamentals in pre-training and fine-tuning to explore microscopic atomic associations. Based on the fact that chemical domain knowledge plays a crucial role in inter-atomic analysis, its enhancement as structural data not only achieves comprehensive information, but also significantly improves data availability and research efficiency. While current knowledge graph research in the field of materials science still mainly relies on open materials databases and patents, papers, etc., there is still a large space for development and research needs in the construction and application of knowledge graph at the elementary level.

\vspace{0.4cm}
\begin{figure}[htb]
  \centering
  \includegraphics[width=5.5in]{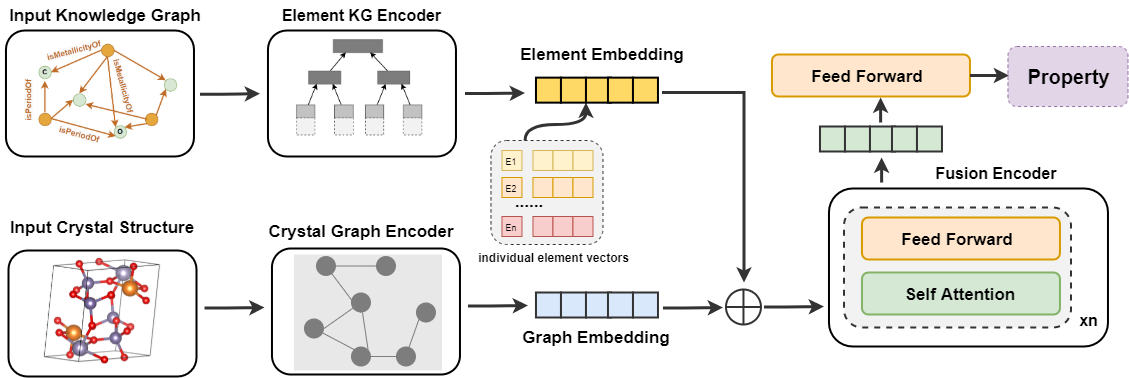}
  \caption{Overview of our adopted methodology ESNet.The complete element embedding is obtained by weighting and summing the individual element vectors output by the Element KG Encoder based on the elemental ratios in the crystal structure.The number of modules n in the Fusion Encoder is set according to the specific task.}
  \label{fig:architecture}
\end{figure} 
\vspace{0.4cm}

\section{Methodology}

In this section, we propose a multi-feature fusion model architecture, ESNet, which integrates elemental properties and crystal structures through a cascaded approach to enhance the accuracy of material property prediction. The ESNet architecture primarily consists of three modules (e.g. Figure~\ref{fig:architecture}): an element KG encoder $\bold M(E)$, which captures the elemental property features $\bold H_e \in \mathbb{R}^{d_1}$, a crystal graph encoder $\bold M(G)$, which extracts the crystal structure features $\bold H_g \in \mathbb{R}^{d_2}$. After features extraction, a fusion encoder $\bold M(F)$, which generates the final joint representation $\bold H_f = \alpha{H_e} + \beta{H_g}$ for predicting the target material properties. Next, we provide a detailed description of each component of the ESNet framework.

\subsection{Element Knowledge Graph Encoder}

Following the ideas of KANO and KCL, we use a similar strategy to construct a knowledge graph and extract elementary feature vectors from the encoded knowledge graph. In the process of implementation, we have made some targeted adjustments and changes. We focus only on the attribute features of the elements covered in the atlas, and further use these filtered attribute features as the features of the crystal structure atoms to achieve the effective enhancement of the features of the crystal structure atoms, thus making the whole chemical element knowledge map more relevant and practical under the new construction.

\begin{figure}[htb]
  \centering
  \includegraphics[width=5in]{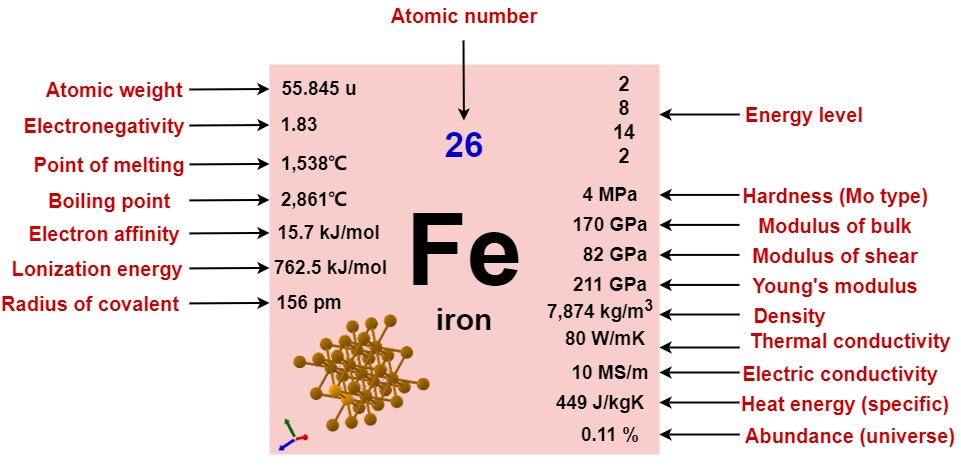}
  \caption{~Examples of Fe element related attributes}\label{Fe}
\end{figure} 
\vspace{0.4cm}

\subsubsection{Element Knowledge Graph Construction}

In our work, all chemical elements and attributes are obtained from the periodic table (\href{https://ptable.com}{https://ptable.com}) , each element contains at least 15 attributes(e.g. Figure~\ref{Fe}). Ternary groups are then constructed based on the obtained elements and corresponding attributes to show the association between elements and attributes. However, for some continuous attributes, it is difficult to perform relationship extraction in the knowledge graph, so we adopt a method to discretise the continuous values, i.e. by dividing the continuous values into different intervals and converting them into discrete labels for effective management in the knowledge graph.

\begin{figure}[htb]
  \centering
  \includegraphics[width=5.5in]{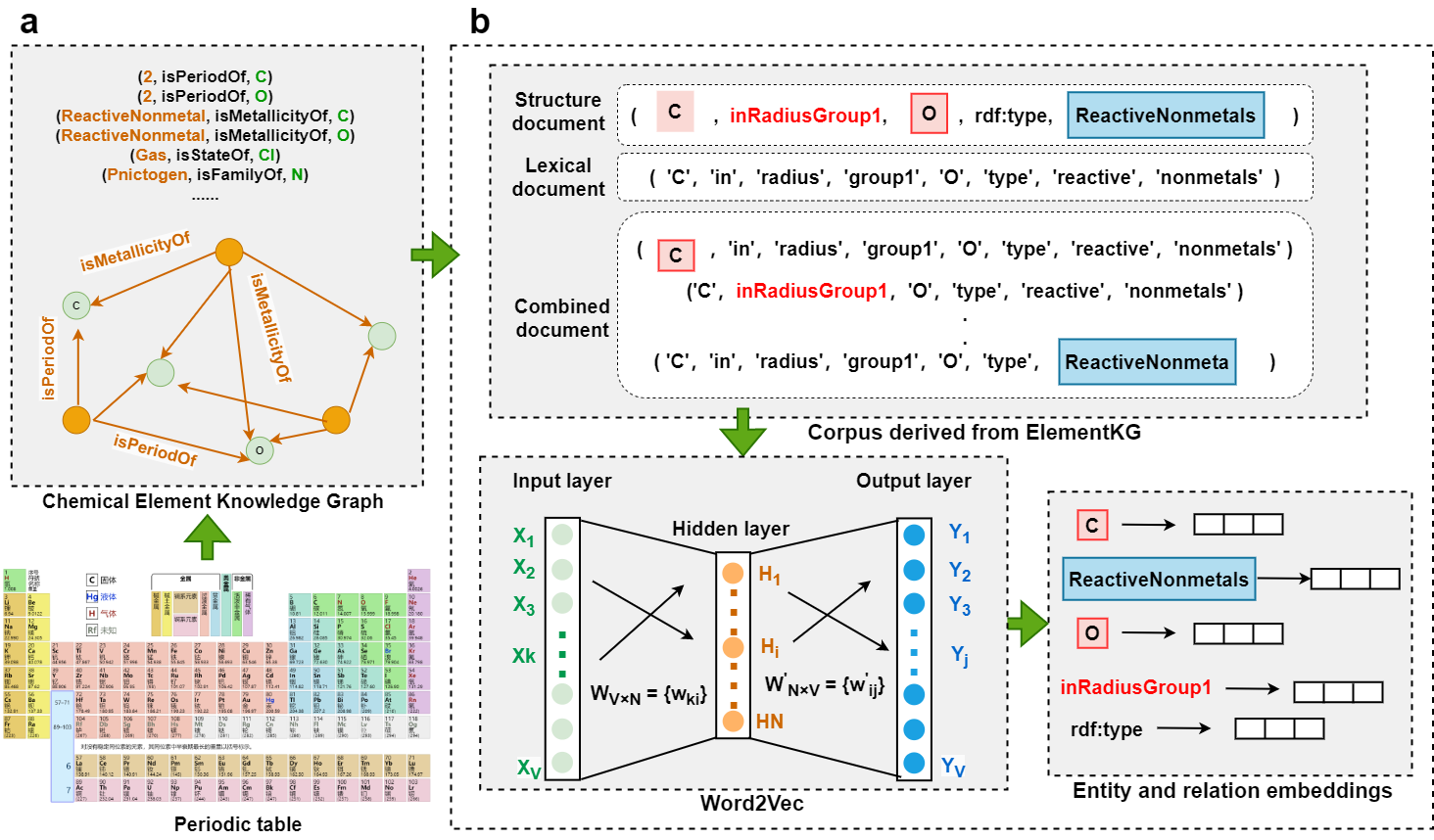}
  \caption{\textbf{Overview of ElementKG Construction And Embedding. 
 a. Construction of the Elemental Knowledge Graph.} We collect element attribute knowledge from the periodic table and construct element triples. \textbf{b.Embedding process for elemental knowledge graph.} Three documents (structural, lexical and combinatorial) are derived from the elemental knowledge graph and merged into a single document for training the word embedding model Word2Vec. This process integrates the elemental attribute knowledge into a unified representation that facilitates the prediction of downstream crystal structure properties.}\label{kg_Embedding}
\end{figure} 
\vspace{0.8cm}

\subsubsection{Element Knowledge Graph Embedding}

During the in-depth study of chemical element knowledge graphs, in order to comprehensively and deeply explore all elemental entities, relationships and semantic information, we adopt a knowledge graph embedding approach based on OWL2Vec\cite{chen2021owl2vec}, a semantic embedding framework specifically designed for the OWL ontology, which encodes the semantics of the ontology through the techniques of random walks on the ontology graph structure and word embedding. OWL2Vec uses Word2Vec\cite{grohe2020word2vec} to train the model to obtain the embedding vectors, but unlike Word2Vec, which focuses on capturing the contextual semantic information of words, OWL2Vec focuses on the OWL ontology, aiming to capture the structural relationships as well as the logical connections of entities and attributes in the ontology, thereby generating low-dimensional vector representations that can be used for machine learning and statistical analysis tasks.

In this paper, an overview of the construction and embedding of chemical element knowledge graphs is shown in Figure ~\ref{kg_Embedding}. We collect basic element knowledge from the periodic table to generate triples, similar to (Density2, isDensityOf, Na). Considering the topological information, semantics and correspondence between entities in the knowledge graph, OWL2Vec converts the OWL ontology into an RDF graph, on which structural, lexical and combinatorial documents are derived, after which the three documents mentioned above are merged, and the merged documents are used to train the word embedding model, Word2Vec, which is finally used to compute the embedding vectors for each target entity.

Since the distribution characteristics of elements in a crystal structure can reflect the physical and chemical properties of the crystal to a certain extent, we have adopted an effective way to reflect this distribution characteristic. We count the embedding vectors of all elements in a crystal structure, and perform weighted superposition of the embedding vectors according to the proportion of elements in the crystal structure, and finally generate the element embedding of the entire crystal structure. For example, for the crystal structure $NaAlSi_3O_8$ , which contains 4 elements Na, Al, Si and O, with a total of 26 atoms, assuming that the retrieved embedding set of all elements is $E(e_1, e_2, e_3, e_4)$, and the proportion of each element in a lattice is $\lambda(\lambda_1, \lambda_2, \lambda_3, \lambda_4)$, then the element embedding of the entire crystal structure is expressed as: $H_e =\lambda_1\times e_1+\lambda_2\times e_2+\lambda_3\times e_3+\lambda_4\times e_4$. We have also confirmed the effectiveness of this method in subsequent experiments.

\subsection{Crystal Graph Encoder}

\begin{figure}[htb]
  \centering
  \includegraphics[width=5.5in]{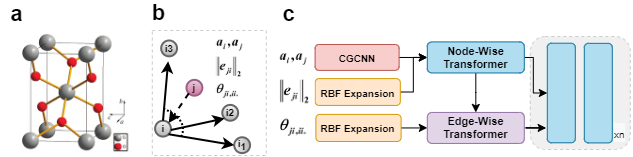}
  \caption{Illustration of the iComFormer pipeline. \textbf{a. Materialcrystal structure.} \textbf{b. Crystal multimodal graph.} \textbf{c. iComFormer architecture}, where elements of the same color belong to the same module.}\label{icomformer}
\end{figure}

ESNet utilizes the crystal transformer iComFormer, which is based on ComFormer(\cite{yan2024complete}), as a graph encoder to capture the node and edge features of crystal graphs. The iComFormer employs SE(3)-invariant crystal graphs for representation learning of crystal structures. Each node $i$ is embedded as an initial node feature ${f}_{i}^{0}$ using CGCNN (\cite{xie2018crystal}). An edge is constructed between nodes $j$ and $i$ if the Euclidean distance ${\parallel{e}_{ji}\parallel}_2$ between them is less than a predefined cutoff radius. Each edge feature is initially mapped as $c/{\parallel{e}_{ji}\parallel}_2$. Then embedded as the initial edge feature $f_{ji}^{e}$ via an RBF kernel. Additionally, the three angles $\{\theta_{ji,ii_1}, \theta_{ji,ii_2}, \theta_{ji,ii_3}\}$ between ${e}_{ji}$ and the lattice representations $\{{e}_{ii_1}, {e}_{ii_2}, {e}_{ii_3}\}$ are mapped to $\{{f}_{ji1}^\theta, {f}_{ji2}^\theta, {f}_{ji3}^\theta\}$. The iComFormer pipeline is illustrated in Figure ~\ref{icomformer}. In this model, the Node-Wise Transformer layer first passes message from a neighboring node $j$ to the center node i using the node features ${f}_{j}^{l}$, ${f}_{i}^{l}$ and edge feature $f_{ji}^{e}$ where $l$ indicates the layer number, and then aggregates all neighboring messages to update ${f}_{i}^{l}$. The Edge-Wise Transformer layer updates edge feature $f_{ji}^{e}$ using angle feature ${f}_{ji1}^\theta$, ${f}_{ji2}^\theta$ and ${f}_{ji3}^\theta$, and edge features of lattice vectors ${f}_{ji1}^{e}$, ${f}_{ji2}^{e}$ and ${f}_{ji3}^{e}$.The final graph feature $H_g$ is obtained by aggregating the updated node and edge features after passing through the iComFormer.

\vspace{0.4cm}

\begin{figure}[htb]
  \centering
  \includegraphics[width=5in]{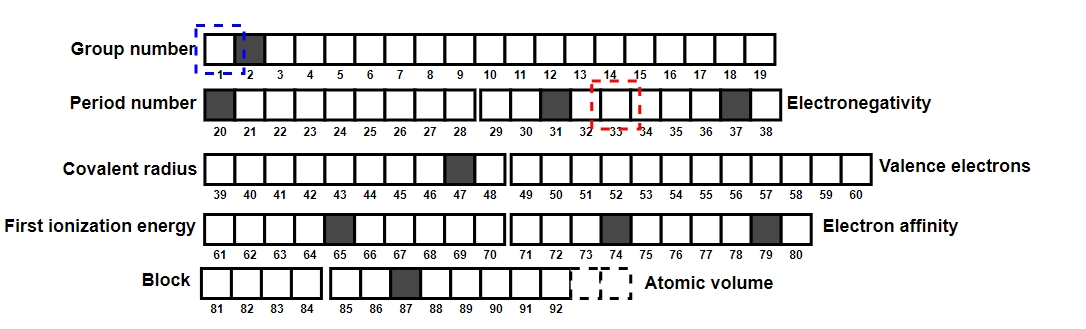}
  \caption{\textbf{"H" atoms vector in CGCNN.} The blue dashed boxes in the figure are used to distinguish other elements from the lanthanide and actinide elements. There are the following issues with the encoding of the hydrogen atom "H": \textbf{a.} The electronegativity is 2.2, so the value in the red dashed box should be set to 1. \textbf{b.} The Valence electrons property is not labeled. \textbf{c.} The "Atomic volume" property should belong to 10 categories, but the data is missing the two values marked by the black dashed box.}\label{fig:one-hot}
\end{figure}

Specifically, we redesigned the atomic encoding in the crystal graph, reducing the original 92-dimensional feature vector encoded by CGCNN(\cite{xie2018crystal}) to 70 dimensions. The motivation behind this redesign is that during the graph construction of the crystal structure, we found that atomic numbers greater than 100 were incorrectly encoded. Additionally, the encoding of other atoms also showed some ambiguities. For example, for the hydrogen atom, the calculated Electronegativity is 2.2, but it is assigned a value of 1 at index 33, as shown in the figure~\ref{fig:one-hot}.

\vspace{0.4cm}

To address these issues, we conducted a statistical analysis using the Mendeleev library on the benchmark Materials Project dataset(\cite{materalsproject}), selecting only categories with complete information, as shown in the table. The encoding method follows the same principles as CGCNN(\cite{xie2018crystal}): for discrete values, the vector is encoded based on the category to which the value belongs; for continuous values, the property value range is uniformly divided into 10 categories, and the vector is encoded accordingly. Additionally, the lanthanide and actinide elements are not only represented by the 8th and 9th periods but also encoded separately, with the first two positions of the vector used to distinguish them. The full list of atom properties as well as their ranges in Table~\ref{atom_property}.

\begin{table}[h!t]
\center
\caption{Properties used in atom features vector}
\begin{tabular}{m{90pt}<{\raggedright}m{90pt}<{\raggedright}m{90pt}<{\raggedright}m{90pt}<{\raggedright}}
\hline\\[-4.5mm]\hline
Property & Unit & Range & \#of categories \\
\hline
Group number & - & 1,2,...,18 & 18\\
Period number & - & 1,2,...,9 & 9\\
Covalent radius & pm & 32-232 & 10\\
Valence electrons & - & 1-17 & 17\\
First ionization energy & eV & 3-65 & 10\\
Block & - & s,p,d,f & 4\\
\hline\\[-4.5mm]\hline
\end{tabular}\label{atom_property}
\end{table}

\begin{figure}[htb]
  \centering
  \includegraphics[width=5.5in]{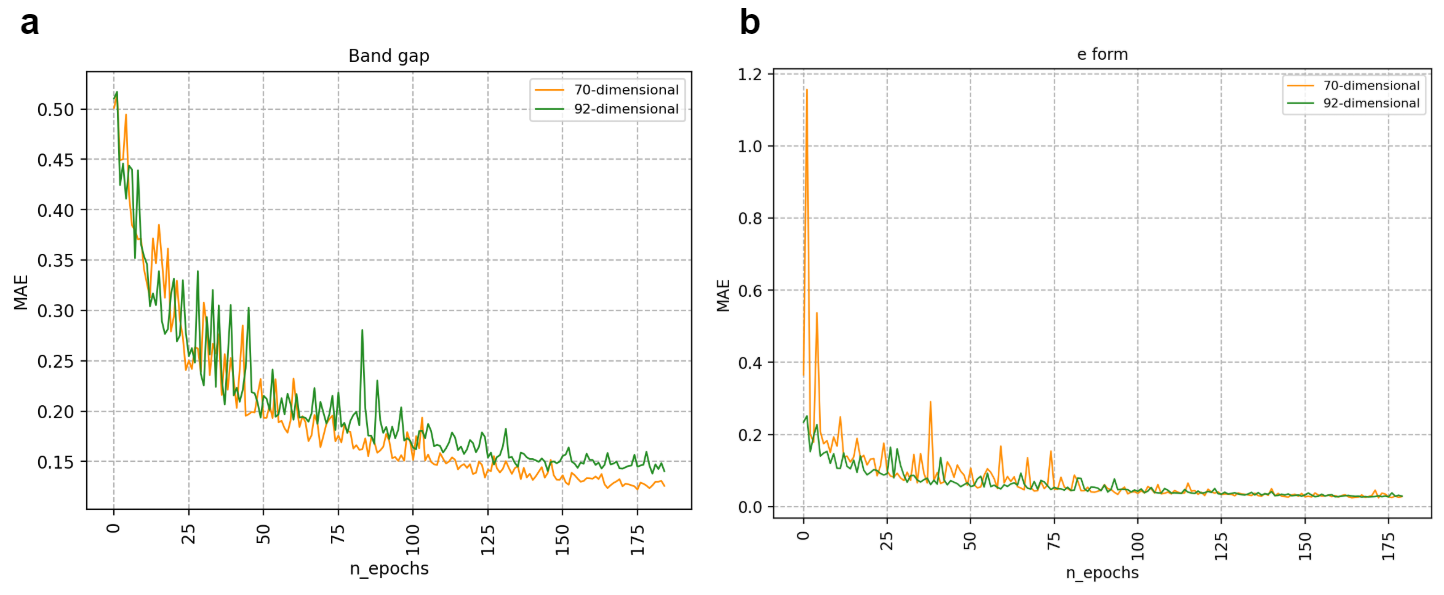}
  \caption{The results on the Materials Project validation set, where both the 70-dimensional and 92-dimensional atomic feature vectors were trained using the same parameters. \textbf{a. The MAE results for the band gap on the validation set}: using L1 loss with Adam optimizer and a starting learning rate of 0.0005. \textbf{b. The MAE results for the formation energy on the validation set}: using MSE loss with Adam optimizer and a starting learning rate of 0.0007.}\label{fig:70-92}
\end{figure}

To verify the impact of the reconstructed 70-dimensional atomic features on the final prediction results, we compared them with the 92-dimensional atomic features used in iComFormer. Keeping the training parameters consistent, we used the band gap and formation energy from the Material Project benchmark dataset as validation metrics. The MAE results show that the 70-dimensional atomic features not only retained all relevant information but also achieved better predictive performance, particularly for band gap prediction, after the same number of training epochs. The results are shown in the figure~\ref{fig:70-92}.

\vspace{0.4cm}

\subsection{Multi-feature Fusion Encoder}
To effectively integrate feature information from different sources, we introduce a multi-feature fusion encoder composed of $n$ Transformer layers. In the fusion encoder, we concatenate the elemental property features $H_e$ (Element Knowledge Graph Encoder output) and the geometric features of the crystal structure $H_g$ (Crystal Graph Encoder output) to create a joint representation: \begin{equation}
    {H_f} = \alpha{H_e} + \beta{H_g}
\end{equation}
Here, $\alpha$ and $\beta$ are weight coefficients used to adjust the contribution of each feature to the final representation across different downstream tasks.

\section{Experiments}
We evaluate the performance of ESNet on the widely-used Materials Project(\cite{materalsproject}) crystal materials benchmark dataset. Through experiments, our proposed ESNet demonstrates excellent performance in predicting material properties such as band gap and formation energy, validating the necessity of incorporating elemental attributes. We follow the experimental setting of ComFormer(\cite{yan2024complete}) and use Mean Absolute Error (MAE) as the evaluation metric. Benchmark methods include CGCNN(\cite{xie2018crystal}), schNet(\cite{schutt2017schnet}), MEGNET(\cite{materalsproject}), GATGNN(\cite{louis2020graph}), ALIGNN(\cite{alignn}), Matformer(\cite{matformer}), PotNet(\cite{lin2023efficient}), CrysMMNet(\cite{das2023crysmmnet}), iComFormer(\cite{yan2024complete}). The the best experimental results highlighted in bold.

\subsection{Experimental results}
The Materials Project-2018.6.1 was first proposed and used by MEGNET(\cite{materalsproject}) collected from The Materials Project(\cite{Commentary}), but the methods are compared by using different random seeds and dataset sizes. To make a fair comparison, (\cite{matformer}) re-train all baseline methods using the same data splits. We follow the experimental settings and data splits of Matformer(\cite{matformer}) and PotNet(\cite{lin2023efficient}). The formation energy and bandgap prediction tasks contain 60000, 5000, and 4239 crystals for training, evaluating, and testing. The evaluation metric used is Mean Absolute Error (MAE). Experimental results on MP are shown in Table~\ref{predict_results}. It can be seen that ESNet shows a significant performance advantage in the bandgap prediction task, while it is roughly comparable to the benchmark results in the formation energy prediction task.

\begin{table}[h!t]
\center
\caption{Comparison on The Materials Project}
\begin{tabular}{m{80pt}<{\raggedright}m{100pt}<{\raggedright}m{100pt}<{\raggedright}}
\hline\\[-4.5mm]\hline
Method & Form.Enery meV/atom & Band Gap  (eV) \\
\hline
CGCNN & 31 & 0.292\\
schNet & 33 & 0.345\\
MEGNET & 30 & 0.307\\
GATGNN & 33 & 0.280\\
ALIGNN & 22 & 0.218\\
Matformer & 21.0 & 0.211\\
PotNet & 18.8 & 0.204\\
CrysMMNet & 20  & 0.197\\
iComFormer & 18.26 & 0.193\\
\hline
SENet & 20.05 & \textbf{0.177}\\
\hline\\[-4.5mm]\hline
\end{tabular}\label{predict_results}
\end{table}

To optimize the performance of the band gap and formation energy prediction tasks, we set different hyperparameters for each task. By adjusting these hyperparameters, we aim to enhance the model's prediction accuracy and ensure that the training process for each task maximizes the model's potential. The specific hyperparameter configurations are shown in the Table~\ref{modelsetting}

\begin{table}[h!t]
\center
\caption{Model settings of ESNet for Material Project}
\begin{tabular}{m{90pt}<{\raggedright}m{90pt}<{\raggedright}m{90pt}<{\raggedright}m{90pt}<{\raggedright}}
\hline\\[-4.5mm]\hline
  & Num. Transformer Layers & Learning rate & Adam \\
\hline
band gap & 8 & 0.0005 & L1 loss\\
formation energy & 4 & 0.0007 & MSE loss\\
\hline\\[-4.5mm]\hline
\end{tabular}\label{modelsetting}
\end{table}

\subsection{Ablation Study}
In this section, we demonstrate the importance of the element knowledge graph encoder in the ESNet framework. Specifically, through ablation experiments, we attempt to directly embed elemental attributes into the crystal structure graph as initial node features. This approach allows us to use iComFormer for predictions without utilizing the element knowledge graph encoder.We conducted experiments on the band gap performance using the MP dataset, and the experimental results are shown in the table. Without using the encoder, the prediction result increased from 0.177 to 0.182.

\begin{table}[h!t]
\center
\caption{Importance of of the Multi-feature Fusion method in the ESNet framework}
\begin{tabular}{m{80pt}<{\raggedright}m{100pt}<{\raggedright}m{100pt}<{\raggedright}}
\hline\\[-4.5mm]\hline
Method & Complete & Test MAE \\
\hline
W/O element knowledge graph encoder & \XSolidBrush & 0.182 \\
\hline
ESNet & \Checkmark & \textbf{0.177}\\
\hline\\[-4.5mm]\hline
\end{tabular}\label{predict_results}
\end{table}

\section{Conclusion and Future Work}

This study presents an innovative material property prediction method by integrating knowledge graph technology with machine learning models. The proposed ESNet framework combines an element knowledge graph encoder with a crystal structure encoder, and leverages a multimodal fusion encoder to capture the intrinsic complex relationships and properties of materials. Compared to existing methods, our model demonstrates a significant improvement in accuracy for predicting band gaps. The performance of our model proves that this approach can effectively enhance prediction accuracy.

The success of ESNet lies in its ability to simultaneously leverage structural and chemical information, thus overcoming the limitations of traditional data-driven models and addressing the lack of chemical prior knowledge. This hybrid approach not only enhances the interpretability of the predictions but also extends the applicability of the model, allowing it to provide effective predictions across a broader chemical space. Our ablation studies further validate the importance of the Element Knowledge Graph Encoder in the ESNet framework and reveal the negative impact on prediction accuracy when this component is absent.

Despite the positive results of our study, several research directions warrant further exploration based on the current findings:
\begin{itemize}
    \item {In the future, we plan to perform more extensive cross-validation on various datasets to assess the generalization ability and robustness of the ESNet model. This work will help us better understand how the model performs across different types of and unknown crystal structures.}
    \item {Currently, the ESNet framework has not been fully tested for predicting shear modulus and bulk modulus. Moving forward, we will focus on predicting these properties, analyzing the key factors affecting their accuracy, improving feature extraction and fusion strategies, and adjusting the model architecture accordingly. Additionally, we plan to collect more samples with precise shear modulus and bulk modulus data to enhance the diversity of the training dataset, thereby improving the model's accuracy and reliability for these properties.}
\end{itemize}

In summary, this study achieves an important breakthrough in the field of material property prediction, successfully integrating knowledge graphs with machine learning and building a promising and powerful tool for materials scientists and engineers.

\bibliography{references}

\end{document}